\documentclass{amsart}
\usepackage{amsmath}
\usepackage[utf8]{inputenc}
\usepackage{graphicx}
\usepackage{textcomp}
\usepackage{xcolor}
\usepackage{cite}
\usepackage[foot]{amsaddr}
\usepackage[utf8]{inputenc}
\usepackage{graphicx}
\usepackage{xcolor}

\title{Grad-CAM++ is Equivalent to \\ Grad-CAM With Positive Gradients}

\author[1]{Miguel Lerma}
\address[1]{Northwestern University, Evanston, USA}
\email[1]{mlerma@math.northwestern.edu}
\author[2]{Mirtha Lucas}
\address[2]{DePaul University, Chicago, USA}
\email[2]{mlucas3@depaul.edu}

\date{\today}

\begin{document}

\maketitle

\begin{abstract}
The Grad-CAM algorithm provides a way to identify what parts of an image
contribute most to the output of a classifier deep network.
The algorithm is simple and widely used for localization of objects in an image, although
some researchers have point out its limitations, and proposed various alternatives.
One of them is Grad-CAM++, that according to its authors
can provide better visual explanations for network predictions,
and does a better job at locating objects even 
for occurrences of multiple object instances in a single image.
Here we show that Grad-CAM++ is practically equivalent to a very simple variation
of Grad-CAM in which gradients are replaced with positive gradients.
\end{abstract}

\section{Introduction}

Artificial Intelligence (AI) has progressed in the last few years
at an accelerated rate, but many AI models behave as black boxes
providing a prediction or solution to a problem without giving any
information about how or why the model arrived to a given conclusion.
This has a negative effect in the trust humans are willing to place in
the output of AI systems. Explainable Artificial Intelligence (XAI)
aims to remediate this problem by providing tools intended to explain
the output of AI models.  Here we look at two of them, Grad-CAM and Grad-CAM++,
how they work, and how they are related. 

We will we start by outlining how Grad-CAM and Grad-CAM++ work. 
Then, we will discuss the methodology behind Grad-CAM++,
and finally we will show how Grad-CAM++ compares to a small variation
of Grad-CAM that we call Grad-CAM${}^+$.

\section{Overview}

Grad-CAM and Grad-CAM++ can be used
on deep convolutional networks whose outputs are differentiable functions.
Their goal is to identify what parts of the network input contribute most
to the output.  In this section we explain how they work.


\subsection{Grad-CAM algorithm}

Here we present two versions of Grad-CAM.  The first version
is the algorithm as described in \cite{selvaraju2017grad}.
The second version, that we name Grad-CAM${}^+$, is a small
variation of Grad-CAM that we have found in some implementations
of the algorithm.

\subsubsection{Grad-CAM}
\begin{figure}[htb]
\begin{center}
\includegraphics[width=3.4in]{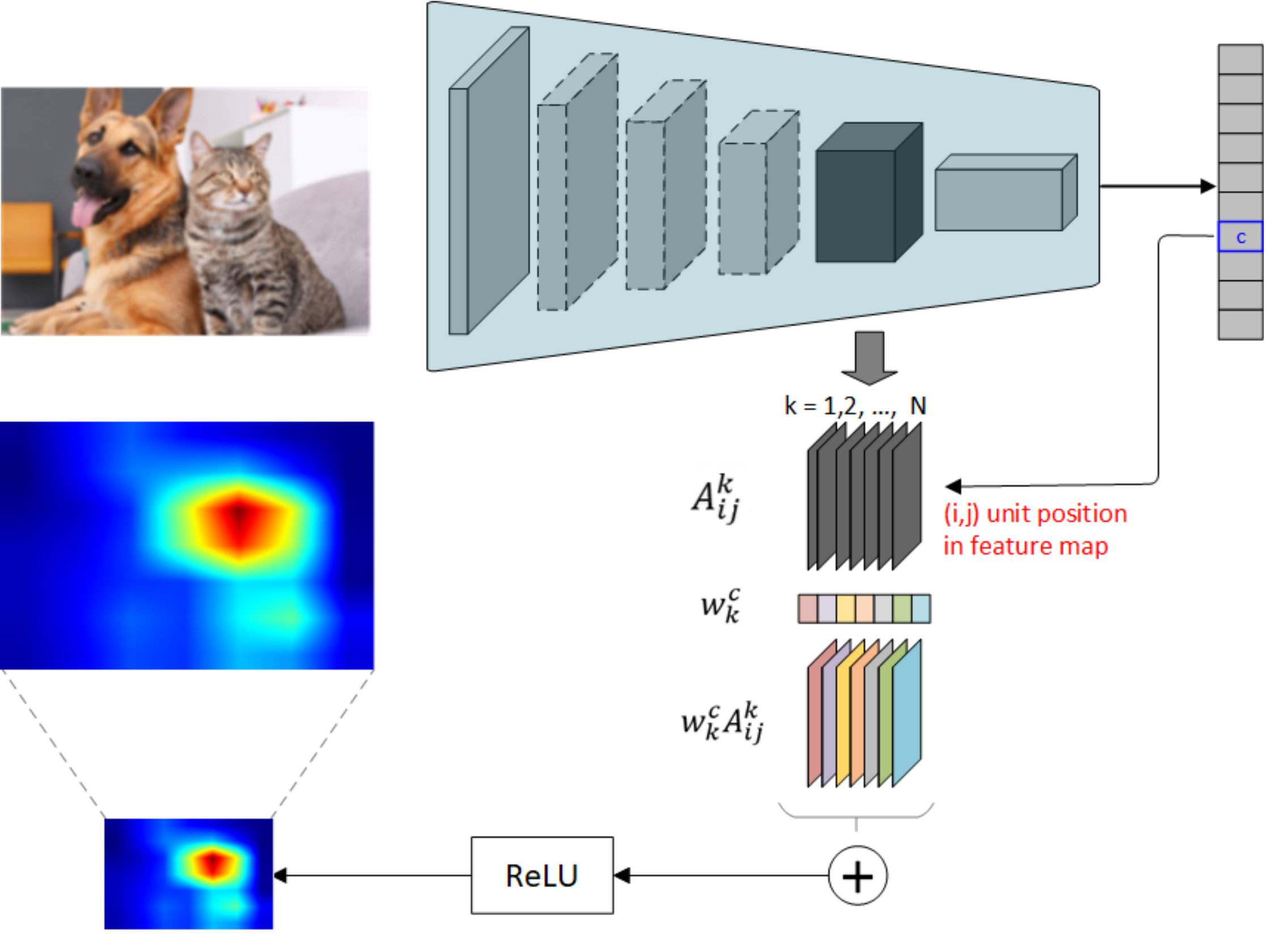}
\caption{Grad-CAM overview.}\label{f:gradcam}
\end{center}
\end{figure}

The Grad-CAM algorithm  works as follows.  First, we must pick a convolutional
layer $A$, which is composed of a number of feature maps or ``channels''
(Figure~\ref{f:gradcam}).  Let $A^k$ be the $k$-th feature map of the picked
layer, and let $A_{ij}^k$ be the activation of the unit in
position $(i,j)$ of the $k$-th feature map.
Then, the localization map,
also called heatmap, saliency map, or attribution map,
is obtained by combining the feature maps of the layer
using weights $w_k^c$ that capture the contribution of the $k$-th
feature map to the output $y^c$ of the network 
corresponding to class~$c$.\footnote{We note that the Grad-CAM 
algorithm as described in \cite{selvaraju2017grad} uses gradients
of pre-softmax scores, we have found implementations in which
gradients of post-softmax scores are used instead---which in
general is easier than using pre-softmax scores e.g. when the model
is given by a third party and pre-softmax scores may not be easy of access.
So in general we will understand that $y^c$ may represent either
the pre-softmax or the post-softmax output of the network,
at the choice of the user.}

In order to compute the weights we pick a class~$c$, and determine how
much the output $y^c$ of the network depends of each unit of the
$k$-th feature map, as measured by the gradient
$\frac{\partial y^c}{\partial A_{ij}^k}$, which can be obtained by
using the backpropagation algorithm.  The gradients are then averaged
thorough the feature map to yield a weight $w_k^c$, as indicated in
equation (\ref{e:gradcam_weights}). Here $Z$ is the size (number of
units) of the feature map.

\begin{equation}\label{e:gradcam_weights}
  w_k^c = \overbrace{\frac{1}{Z} \sum_{i}\sum_{j}}^{\text{global average pooling}}
  \hskip -30pt
  \underbrace{\frac{\partial y^c}{\partial A_{ij}^k}}_{\text{gradients via backprop}}
  \,.
\end{equation}

The next step consists of combining the feature maps $A^k$ using the
weights computed above, as shown in equation (\ref{e:heatmap}).  Note
that the combination is also followed by a Rectified Linear function
$\text{ReLU}(x) = \text{max}(x,0)$,
because we are interested only in the features that have a positive
influence on the class of interest. The result $L_{\text{Grad-CAM}}^c$
is called \emph{class-discriminative localization map} by the authors. 
It can be interpreted as a coarse heatmap of the same size as the chosen
convolutional feature map.

\begin{equation}\label{e:heatmap}
  L^c =
  \text{ReLU} \underbrace{\Biggl(\sum_k w_k^c A^k\Biggr)}_{\text{linear combination}}
  \,.
\end{equation}

\begin{figure}[htb]
\begin{center}
\includegraphics[height=1in]{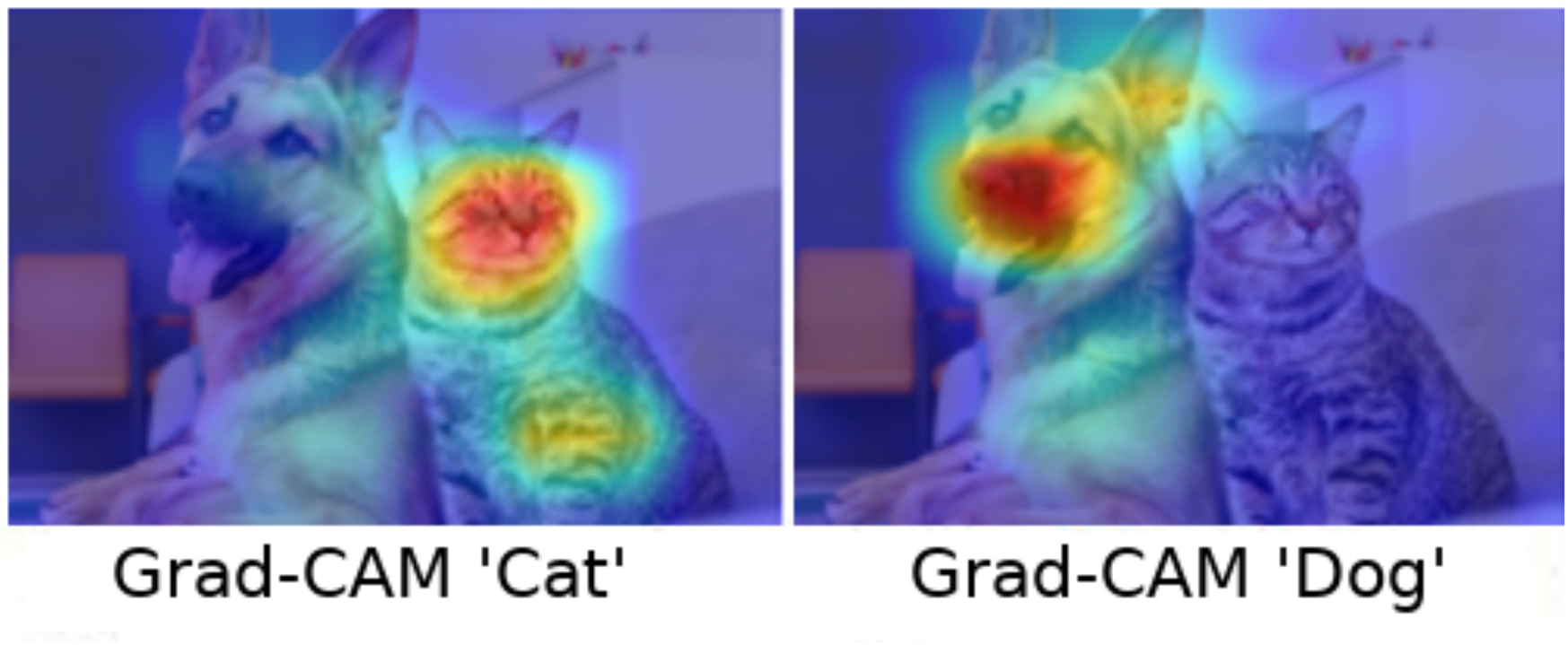}
\caption{Grad-CAM locating a cat and a dog. .}\label{f:catdog}
\end{center}
\end{figure}

After the heatmap has been produced, it can be normalized and
upsampled via bilinear interpolation to the size of the original image,
and overlapped with it to highlight the areas of the input image that
contribute to the network output corresponding to the chosen class.
Figure~\ref{f:catdog} shows the resulting heatmap (after applying a colormap) 
obtained in the same figure for classes `cat' and `dog' respectively.
The red color indicates the areas in which the heatmaps have a higher intensity, 
which are expected to coincide with the location of the objects corresponding
to the classes picked.

The method is very general, and can be applied to any (differentiable)
network outputs.

\subsubsection{Grad-CAM${}^+$}
In some implementations of Grad-CAM we have found that in the computation
of the weights only terms with \emph{positive} gradients are used, i.e.,
the weights are computed using the following formula:
\begin{equation}\label{e:gradcam_weights1}
  w_k^c = \frac{1}{Z} \sum_{i}\sum_{j}
  \text{ReLU}\Bigl(\frac{\partial y^c}{\partial A_{ij}^k}\Bigr)
  \,.
\end{equation}
This can be justified by the
intuition that negative gradients correspond to units where
features from a class different to the chosen class are present
(say an area showing a `cat' when trying to locate a `dog' in an image). 
Although in the implementations this version of the Grad-CAM algorithm
with positive gradients is still named ``Grad-CAM,''
for clarity here we denote it ``Grad-CAM${}^+$.''


\subsection{Grad-CAM++ algorithm}
Grad-CAM++ has been introduced with the purpose of providing a better localization
than Grad-CAM \cite{chattopadhyay2018gradplus}.
When instances of an object appear in various places, or produce
footprints in different feature maps, a plain average of the gradients at each feature
map may not provide a good localization of the regions of interest because the units
of each feature map may have different ``importance'' not fully captured by the gradient alone.
The solution proposed in \cite{chattopadhyay2018gradplus}
is to replace the plain average of gradients at each feature map 
with a weighted average:
\begin{equation}\label{e:weights}
  w_k^c = \sum_{i}\sum_{j} \alpha_{ij}^{kc} \cdot \text{ReLU}\Bigl( \frac{\partial y^c}{\partial A_{ij}^k} \Bigr)
  \,,
\end{equation}
where the $\alpha_{ij}^{kc}$ measure the importance of each individual unit of a feature map.
Note also that the gradients are fed to a $\text{ReLU}$ function, so only positive
gradients are considered at this step.

In order to determine the values of the alphas the following equation is posed:
\begin{equation}\label{e:scoresfrompooling}
Y^c = \sum_k w_k^c \cdot \sum_{i,j} A_{ij}^k \,,
\end{equation}
i.e., the global pooling of the feature maps of the layer, combined using the weights $w_k^c$, should produce the class scores $Y^c$.
Next, plugging (\ref{e:weights}) in (\ref{e:scoresfrompooling}) we get:
\begin{equation}\label{e:alphasequation}
    Y^c =
    \sum_k \Bigl( \Bigl\{ \sum_{a,b} \alpha_{ab}^{kc} \cdot \text{ReLU}\Bigl( \frac{\partial Y^c}{\partial A_{ab}^k} \Bigr) \Bigr\}
    \Bigl[ \sum_{i,j} A_{ij}^k \Bigr] \Bigr) 
    \,.
\end{equation}

In order to isolate the alphas, partial derivatives w.r.t. $A_{ij}^k$ are computed 
on both sides of~(\ref{e:alphasequation})
(without the ReLU function), obtaining (equation (8) in the paper):
\begin{equation}\label{e:firststep}
    \frac{\partial Y^c}{\partial A_{jk}^k} = \sum_{a,b} \alpha_{ab}^{kc} \cdot \frac{\partial Y^c}{\partial A_{ab}^k} + 
    \sum_{a,b} A_{ab}^k \Bigl\{ \alpha_{ij}^{kc} 
    \cdot \frac{\partial^2 Y^c}{(\partial A_{jk}^k)^2} \Bigr\}
    \,.
\end{equation}

Taking partial derivative w.r.t. $A_{ij}^k$ again and isolating 
$\alpha_{ij}^{kc}$ the following equation is obtained:
\begin{equation}\label{e:alphaformula}
\alpha_{ij}^{kc} =
\frac{\frac{\partial^2 Y^c}{(\partial A_{ij}^k)^2}}
{2\frac{\partial^2 Y^c}{(\partial A_{ij}^k)^2} + \sum_{ab} A_{ab}^k \{ \frac{\partial^3 Y^c}{(\partial A_{ij}^k)^3} \}} 
\,.
\end{equation}
The final expression for the alphas involve second and third order partial derivatives.
Assuming that the score $Y^c$ is an exponential of the pre-softmax 
output of the network $S^c$, i.e., $Y^c = \exp(S^c)$, the expression yielding the alphas
becomes:
\begin{equation}\label{e:finalalpha}
\alpha_{ij}^{kc} =
\frac{\Bigl(\frac{\partial S^c}{\partial A_{ij}^k}\Bigr)^2}
{2\Bigl(\frac{\partial S^c}{\partial A_{ij}^k}\Bigr)^2 + \sum_{ab} A_{ab}^k  \Bigl(\frac{\partial S^c}{\partial A_{ij}^k}\Bigr)^3 } 
\,,
\end{equation}
which does not involve high order partial derivatives.
This final equation (\ref{e:finalalpha}) is the one used in actual implementations.
Note that if $\frac{\partial S^c}{\partial A_{ij}^k} = 0$ then the right hand side of (\ref{e:finalalpha})
becomes $0/0$, which is  undefined.  In this case $\alpha_{ij}^{kc}$ is assigned value zero.
After the alphas are computed the weights are obtained using equation~(\ref{e:weights}), and then the salience map using 
equation~(\ref{e:heatmap}).\footnote{Note that it does not
matter if we use $y^c=\exp(S^c)$, hence
$\frac{\partial y^c}{\partial A_{ij}^k} = \exp(S^c) \frac{\partial S^c}{\partial A_{ij}^k}$, or $y^c=S^c$, $\frac{\partial y^c}{\partial A_{ij}^k} = 
\frac{\partial S^c}{\partial A_{ij}^k}$. The heatmaps obtained
one way or the other differ by the constant factor $\exp(S^c)$, which
will have no effect after min-max normalization of the heatmaps.
It fact, because of the rapid grow of the exponential function,
which may lead to numerical instability, the choice $y^c=S^c$ is
arguably better than $y^c=\exp(S^c)$.}

\section{Discussion}

In this section we discuss several issues we found in the algorithm for Grad-CAM++
and its derivation.

\subsection{The math is unclear}
In this section we focus on mathematical issues.

\subsubsection{Equation (\ref{e:alphasequation}) is underdetermined.}
Equation (\ref{e:alphasequation}) is a system of linear equations
with more unknowns (the~$\alpha_{ij}^{kc}$) than equations (just one per class),
which makes it underdetermined.  More specifically, equation (\ref{e:alphasequation})
can be written:
\begin{equation}\label{e:alphasequation2}
    \sum_{k,a,b} C_{ab}^k \cdot \alpha_{ab}^{kc} = Y^c
    \,,
\end{equation}
where $C_{ab}^k = \text{ReLU}\Bigl( \frac{\partial Y^c}{\partial A_{ab}^k} \Bigr) \Bigl[ \sum_{i,j} A_{ij}^k \Bigr]$, or 
$C_{ab}^k = \frac{\partial Y^c}{\partial A_{ab}^k} \Bigl[ \sum_{i,j} A_{ij}^k \Bigr]$
if we remove the $\text{ReLU}$.
Note that (\ref{e:alphasequation2}) has infinitely many solutions. 
A~general set of solutions can be written like this:
\begin{equation}
    \alpha_{ab}^{kc} = 
    \beta_{ab}^{kc} \cdot Y^c \cdot \Bigl( \sum_{k,a,b} C_{ab}^k \cdot \beta_{ab}^{kc} \Bigr)^{-1}
    \,,
\end{equation}
where $\beta_{ab}^{kc}$ are arbitrary numbers constrained only by 
the condition $\sum_{k,a,b} C_{ab}^k \cdot \beta_{ab}^{kc} \neq 0$.
Consequently, isolating the $\alpha_{ij}^{kc}$
to get a single solution like in equation (\ref{e:alphaformula})
is impossible without adding additional restrictions.

\subsubsection{The alphas are treated as constants, however they may not be.}
In the derivation of equation 
(\ref{e:firststep}) the $\alpha_{ij}^{kc}$ 
are treated as constants, but that cannot be correct because they
depend on the $A_{ij}^{k}$.  In fact the method used to isolate the alphas
is equivalent to ``solving'' the equation $\alpha x = x^2$ in $\alpha$
by differentiating on both sides w.r.t $x$ and concluding that $\alpha = 2x$,
which is obviously incorrect.
The actual result of differentiating on both sides of this equation
should read $\frac{d\alpha}{d x} x + \alpha = 2x$, which is correct, but does not
help isolate the $\alpha$, it only complicates unnecessarily the original
equation.

\subsubsection{Issues with the computation of partial derivatives.}
Even if we assume that the $\alpha_{ij}^{kc}$ are constant,
taking partial derivative of~(\ref{e:alphasequation})
w.r.t. $A_{ij}^k$ does not yield (\ref{e:firststep}).
The computation (assuming that the $\alpha_{ij}^{kc}$ are constant) 
is shown in the appendix---note the extra summation and the partial cross-derivatives.

\subsubsection{The alphas obtained may not satisfy their defining equation.}
Another aspect that interferes with the process of solving equation
(\ref{e:alphasequation})
is that second degree derivatives kill linearities, so if after computing the alphas
we replace $Y^c$ with $Y^c$ plus a linear function of the $A_{ij}^k$, i.e.,
$Y^c \to Y^c + \sum_{i,j,k}\lambda_{ijk} A_{ij}^k + C$, where $\lambda_{ijk}$ 
and $C$ are constants, the process would lead to the same values
for the~$\alpha_{ij}^{kc}$, even though they cannot be solutions to the original
and new equations simultaneously.

\subsubsection{About the assumption $Y^c = \exp(S^c)$.}
The assumption that the score $Y^c$ is an exponential of the pre-softmax 
output of the network $S^c$ is based on the fact that the $Y^c$ can be any 
(increasing) smooth function of $S^c$.\footnote{A smooth function
is a function that is differentiable up to some desirable order.} 
But if that is the case then the
substitution $Y^c = \exp(\lambda S^c)$, where $\lambda$ is any positive constant,
would also be legitimate since $f_{\lambda}(x) = e^{\lambda x}$ is also smooth.
However this would produce a different final formula for
the alphas, namely:
\begin{equation}\label{e:finalalpha3}
\alpha_{ij}^{kc} =
\frac{\Bigl(\frac{\partial S^c}{\partial A_{ij}^k}\Bigr)^2}
{2\Bigl(\frac{\partial S^c}{\partial A_{ij}^k}\Bigr)^2 +
\lambda \sum_{ab} A_{ab}^k  \Bigl(\frac{\partial S^c}{\partial A_{ij}^k}\Bigr)^3 } 
\,.
\end{equation}
In principle there does not seem to be any particular reason to choose
$\lambda=1$ rather than any other value, and different values of $\lambda$ 
will lead to different values for the alphas.

Leaving aside the question of the derivation of the formula for the alphas,
we next discuss the formula actually used in the implementations to compute the alphas.

\subsection{Formula used in actual implementations}
The formula for $\alpha_{ij}^{kc}$ actually used in most implementations
of Grad-CAM++ we have found is (\ref{e:finalalpha}). 
Note that if we divide
numerator and denominator by $(\partial S^c/\partial A_{ij}^k)^2$ we get
\begin{equation}\label{e:finalalpha2}
    \alpha_{ij}^{kc} =  \frac{1}{2 + \sum_{ab} A_{ab}^k  \Bigl(\frac{\partial S^c}{\partial A_{ij}^k}\Bigr)}
    \,,
\end{equation}
which is mathematically equivalent to (\ref{e:finalalpha}), except 
when the gradients are zero, in which case $\alpha_{ij}^{kc} = 0$.

Although the formula does not contain second and third powers anymore
(which reduces the risk of under or overflow) the expression is still numerically
unstable, because nothing prevents the denominator from getting close to zero.
In our experiments we have observed that this in fact happens occasionally.
Also, if the second term in the denominator of (\ref{e:finalalpha2})
remains small compared to $2$, the alphas will be approximately constant.
Next, we discuss this two issues.

\subsection{In practice the alphas are nearly constant}
\begin{figure}[htb]
\begin{center}
\includegraphics[height=2in]{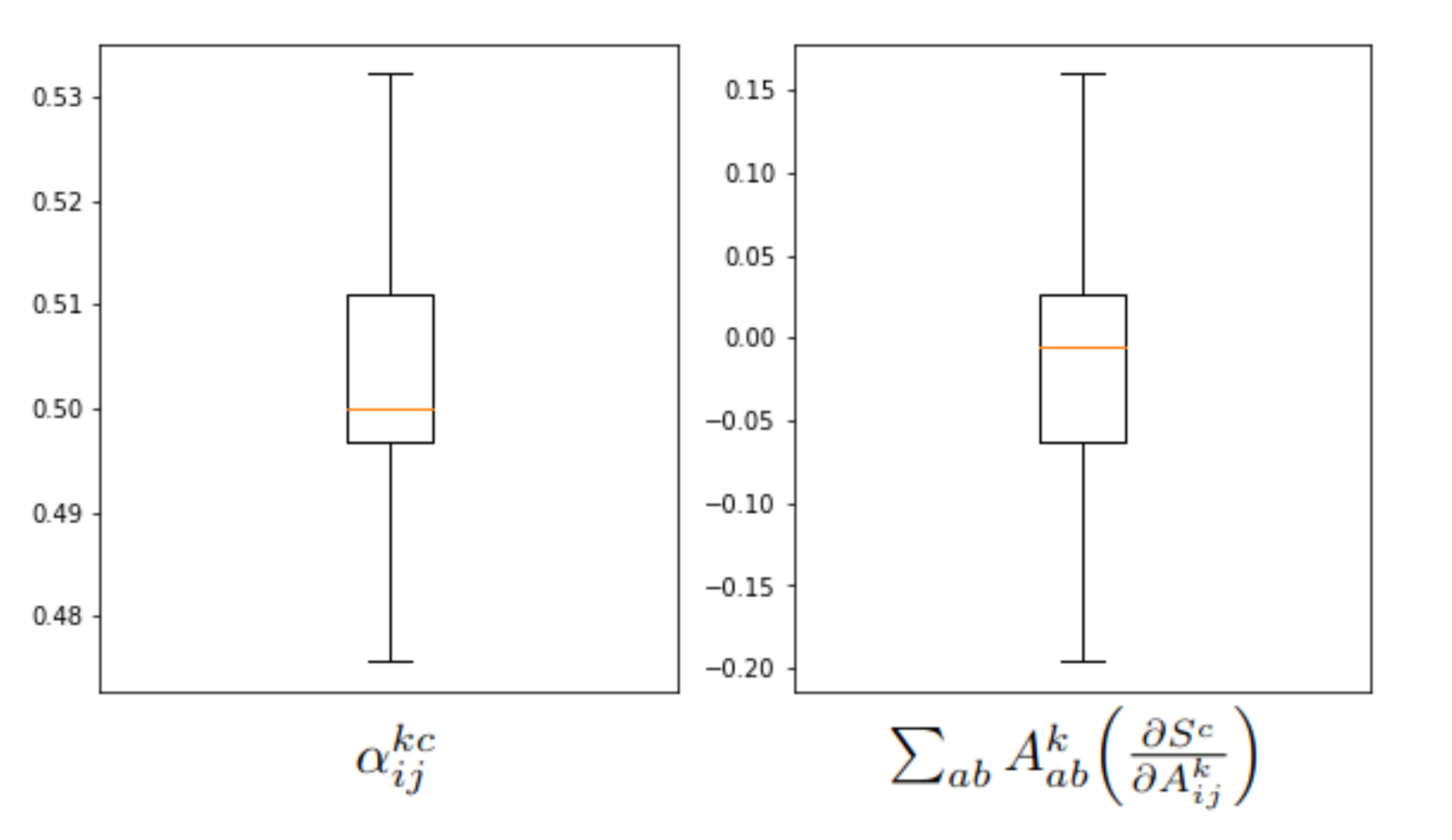}
\caption{Example of distributions of non-zero $\alpha_{ij}^{kc}$
and $\sum_{ab} A_{ij}^k \Bigl( \frac{\partial S^c}{\partial A_{ab}^k} \Bigr)$
for a single image.}\label{f:alphaboxplot}
\end{center}
\end{figure}

We have observed that the absolute value of the second term
in the denominator of (\ref{e:finalalpha2}) is usually small compared to 2,
and consequently the alphas tend to take values around $1/2$.
As an illustration of this point Figure~\ref{f:alphaboxplot}
shows the boxplots for the distribution of non-zero values 
(after removal of outliers) of
$a_{ij}^{kc}$ and $\sum_{ab} A_{ab}^k\Bigl(\frac{\partial S^c}{\partial A_{ij}^k}\Bigr)$
in the last convolutional layer (block5\_conv3)
of a VGG16 \cite{simonayan2015} fed with a sample image.  In this particular
example most values of $\sum_{ab} A_{ab}^k\Bigl(\frac{\partial S^c}{\partial A_{ij}^k}\Bigr)$
lie in the interval $(-0.2,0.15)$, and
the resulting values for $\alpha_{ij}^{kc}$ range between 0.48 and 0.53 approximately.

\begin{figure}[htb]
\begin{center}
\includegraphics[height=2.2in]{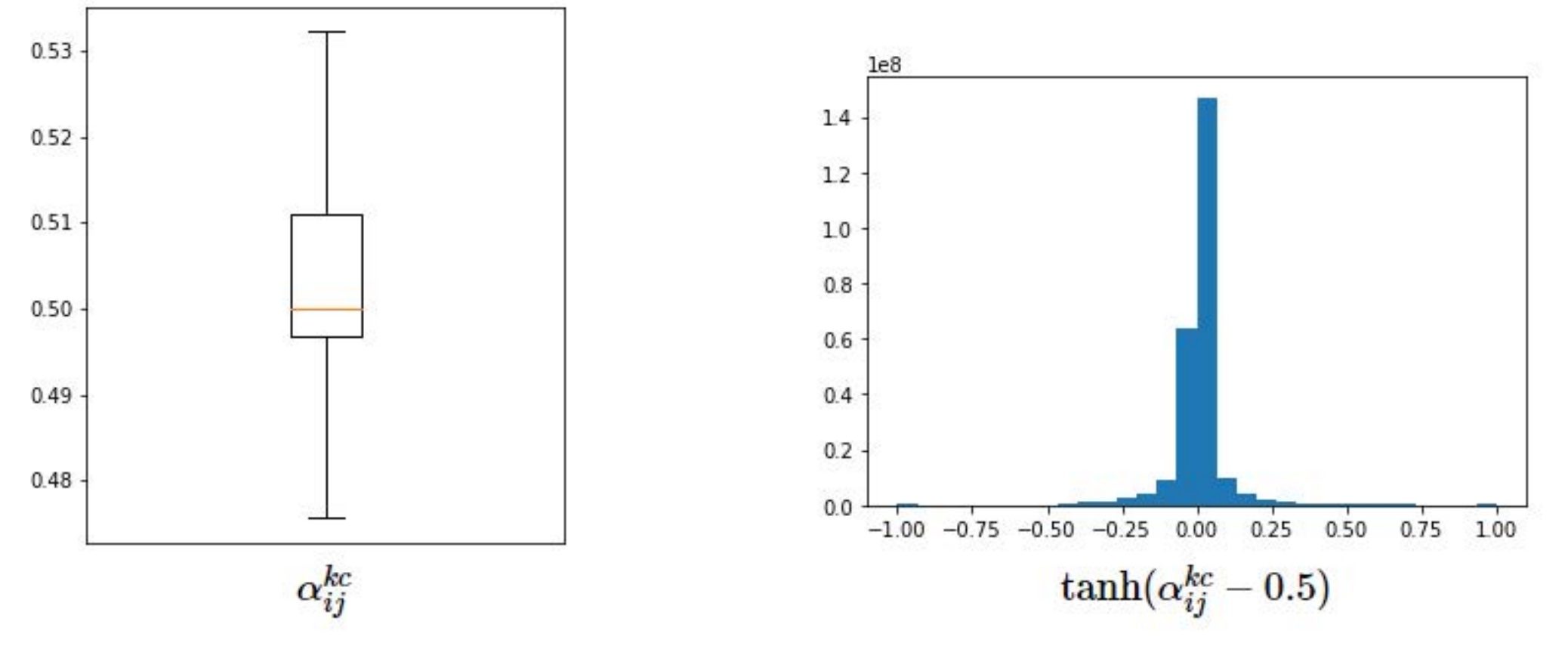}
\caption{Distribution of (non-zero) $\alpha_{ij}^{kc}$ across
10,000 images from ImageNetV2. On the left a boxplot
of the values of $\alpha_{ij}^{kc}$ (with outliers removed) is shown.
On the right there is a histogram of the values of
$\tanh(\alpha_{ij}^{kc} - 0.5)$.}\label{f:alphaboxplot-imagenetv2}
\end{center}
\end{figure}

This happened consistently for all images we tried,
but in order to get a more solid result we plotted also 
the distribution of the alphas obtained after feeding the 
VGG16 with 10,000 images from ImageNetV2 MatchedFrequency \cite{imagenetv2}.
The result is shown in Figure~\ref{f:alphaboxplot-imagenetv2}.
For the boxplot we have considered non-zero values of $\alpha_{ij}^{kc}$
only, and removed outliers. 
The reason to consider only non-zero
values for the alphas is that
$\alpha_{ij}^{kc}$ is set to zero precisely when the gradients
vanish, and in that case the values of $\alpha_{ij}^{kc}$ do not play any role
because if $\frac{\partial S^c}{\partial A_{ij}^k} = 0$ then
$\alpha_{ij}^{kc} \cdot \text{ReLU}\Bigl( \frac{\partial S^c}{\partial A_{ij}^k} \Bigr) = 0$
regardless of the value of $\alpha_{ij}^{kc}$.
For the histogram on the right we have included
again all non-zero values of alpha, but without removing outliers,
centered at $0.5$ and squeezed with an hyperbolic tangent, i.e., 
$\tanh(\alpha_{ij}^{kc} - 0.5)$.  The $\tanh$-squeezing
allows to display all the frequency bars within the interval $[-1,1]$
even though the outliers may take values very far away from $0.5$.

\begin{figure}[htb]
\begin{center}
\includegraphics[height=2.5in]{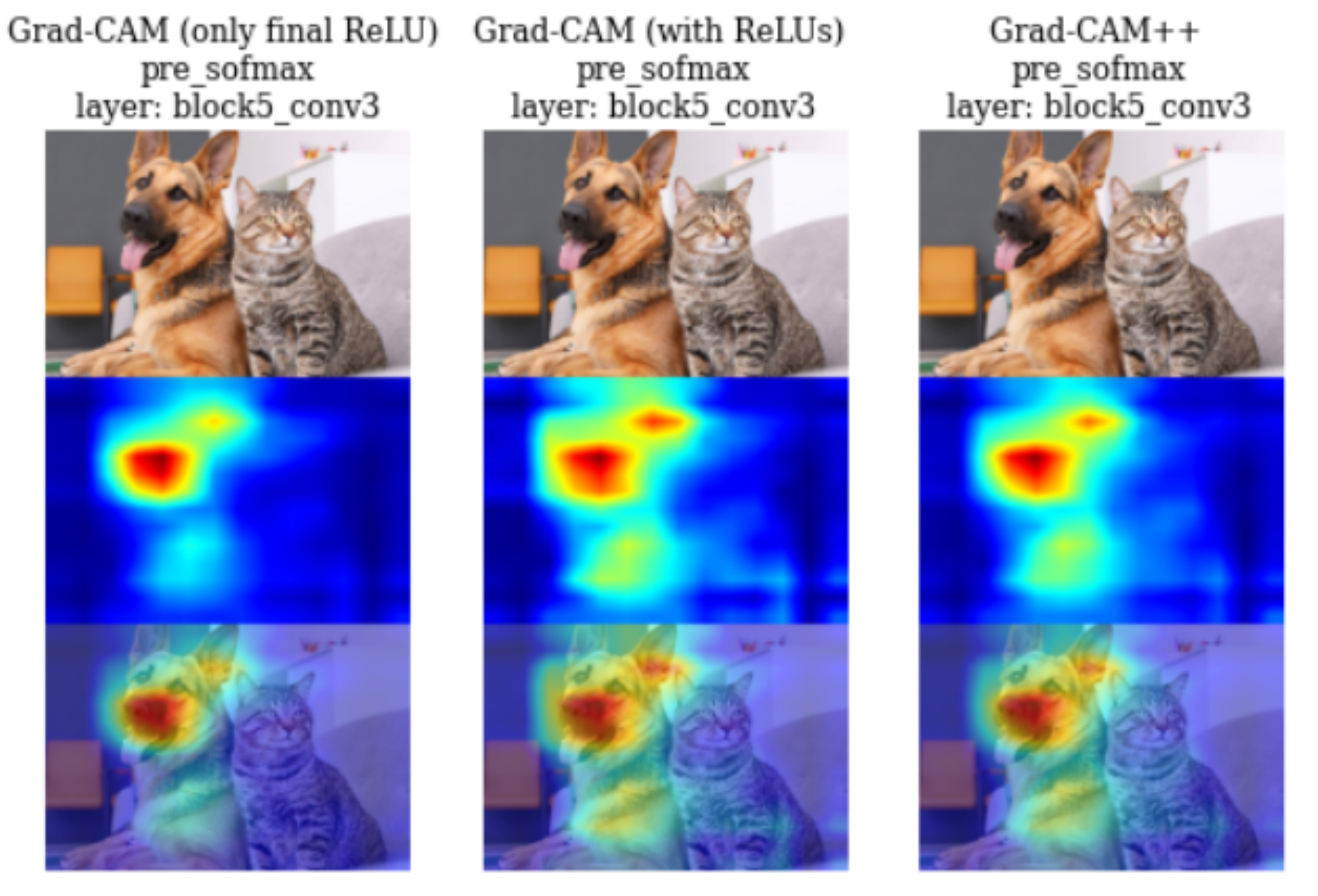}
\caption{Heatmaps produced by Grad-CAM with only the final ReLU (left),
Grad-CAM with positive gradients, i.e. Grad-CAM${}^+$ (center),
and Grad-CAM++ (right) respectively. 
Note that the last two heatmaps are very similar.}\label{f:dog-and-cat}
\end{center}
\end{figure}

Consequently, we have $\alpha_{ij}^{kc} \approx \frac{1}{2}$, 
and
\begin{equation}\label{weights2}
  w_k^c \approx \frac{1}{2} \sum_{i}\sum_{j} \text{ReLU}\Bigl( \frac{\partial y^c}{\partial A_{ij}^k} \Bigr)
  \,,
\end{equation}
hence, the weights computed are (except for a multiplicative constant) 
approximately the same as
the weights computed for Grad-CAM${}^+$,
i.e., Grad-CAM replacing gradients with $\text{ReLU}$
of gradients. As a consequence, in most cases we expect
the heatmap produced by Grad-CAM++ to be practically the
same as the one obtained using Grad-CAM${}^+$ given by equation~(\ref{e:gradcam_weights1}).
Figure~\ref{f:dog-and-cat} is an illustrative example in which Grad-CAM++
and Grad-CAM with positive gradients (Grad-CAM${}^+$) produce nearly identical heatmaps.

\subsection{The computation of the alphas is numerically unstable}
Even though the alphas tend to take values around $0.5$ we also noticed
that for a few outliers $\alpha_{ij}^{kc}$ reached values
as large as $-83.78$ and $42.94$ for the single image we tried.
On the whole ImageNetV2 the extreme values were
$-1398101.4$ and  $559240.56$ respectively,
which confirmed the fact that the computation of the alphas using
equation (\ref{e:finalalpha2}) is
numerically unstable.  Furthermore, we cannot think of any
theoretical justification to assign extremely large values to the ``pixel-importance''
(as measured by $a_{ij}^{kc}$) when the value of 
$\sum_{ab} A_{ab}^k \Bigl( \frac{\partial S^c}{\partial A_{ij}^k} \Bigr)$
happens to be close to $-2$.

\section{Empirical tests}

The previous section contains mainly theoretical considerations,
but the fact remains that the literature shows Grad-CAM++ usually
performing better than Grad-CAM. Here we examine a possible explanation.
Our hypothesis is that the main factor that makes Grad-CAM++ better than
the original version of Grad-CAM is not the Grad-CAM++ algorithm per se,
but the fact that in the computation of the weights only positive
gradients are used.  A result that can be presented in support of this hypothesis
is section 7 of \cite{chattopadhyay2018gradplus}, which
shows that the performance of Grad-CAM++ drops significantly when
the requirement of using only positive gradients is dropped.
We will see that using only positive gradients in the computation of
the weights is not only necessary, it is also sufficient in the sense that
(in most cases) Grad-CAM++ does not do significantly better than 
Grad-CAM with positive gradients (Grad-CAM${}^+$),
i.e., the version of Grad-CAM in which weights are 
computed using equation~(\ref{e:gradcam_weights1}).
Note that the different coefficients in front of the sum are inconsequential
because the heatmaps are ultimately min-max normalized.

To compare the performances of the three methods we use a metric
loosely inspired in the average drop, increase in confidence,
and win-percentage metrics described in
\cite{chattopadhyay2018gradplus}, with a difference: rather than
averaging percentages with an arithmetic mean we will average
proportions using geometric mean.  The reason for this choice is that,
in general, adding, subtracting, and averaging percentages is not a good
idea and can lead to wrong results.

The first step is to produce
explanation maps, defined as the Hadamar (point-wise) product
of images and their (min-max normalized) corresponding heatmaps
generated by the attribution methods:
\begin{equation}
    E^c = L^c \odot I
    \,.
\end{equation}
The explanation map $E^c$ can be interpreted as the original image $I$
in which the areas with least contribution to the output of the network
have been obscured.
Hence, if we feed the explanation map $E^c$ to the network, we expect
that the output of the network will be larger when the explanation
map correctly captures the relevant areas of the image compared to
the output obtained if the relevant areas are poorly captured.\footnote{Note that
regardless of whether explanation maps are the best way to evaluate attribution
methods, they can still be used for comparison purposes since methods producing
similar heatmaps will also produce similar explanation maps.}

Then, the relative performance of two attribution
methods can be assessed by comparing the network outputs after feeding
the network with each of the corresponding explanation maps. More specifically, 
let $I_i$ be the $i$th image from the dataset,
let $E_i^c $, $E'^c_i$ be explanation maps produced
for $I_i$ using attribution methods $M$ and $M'$, and let $O^c_i$,
$O'^c_i$ be the network outputs (after softmax) obtained when
feeding the network with $E_i^c $, $E'^c_i$. 
We call the quotient $O'^c_i/O^c_i$
\emph{relative performance} of $M'$ vs $M$ for image $I_i$.
Note that the relative performance will be larger than 1 if the heatmap
produced by method $M'$ captures the relevant areas of $I_i$
better than the heatmap produced by method $M$ does,
otherwise $O'^c_i/O^c_i$ will be less than 1.  The relative performance
of methods $M$ and $M'$ across a dataset
can be measured as the geometric mean of the
product of $O'^c_i/O^c_i$ across the given dataset:
\begin{equation}
    \text{relative performance (across a dataset)} = \sqrt[n]{\prod_{i=1}^n \frac{O'^c_i}{O^c_i}}
    \,.
\end{equation}
Note that we average the ratio $O'^c_i/O^c_i$ using a geometric rather than
arithmetic mean.  This choice is consistent with the use of the geometric mean
in fields in which amounts are compared using ratios rather than differences
(e.g. growth rates, financial indices, etc.).  
For some statistics we also use the log relative performance
per image defined as $\log(O'^c_i/O^c_i) = \log{O'^c_i} - \log{O^c_i}$.
Note that the relative performance across a dataset is the exponential
of the arithmetic mean of the log relative performance across that dataset:
\begin{equation}
\text{relative performance} = 
\exp\Bigl\{\frac{1}{n} \sum_{i=1}^n \log\Bigl(\frac{O'^c_i}{O^c_i}\Bigr)\Bigr\}
\,.
\end{equation}

The testing dataset used is ImageNetV2 MatchedFrequency, which contains 10,000
images from 1,000 categories~\cite{imagenetv2}. In order to
separate network performance from attribution method performance, we restrict
the statistics to a subset of 4219 images for which the network assigns 
a (post-softmax) score of more than $0.5$ to the right class.

After applying the relative performance metric to each pair of algorithms of 
(original) Grad-CAM, Grad-CAM${}^+$, and Grad-CAM++ 
across the dataset, we get
the results shown in Table~\ref{t:relperformance}.
The two column relative performance shows results obtained
with explanation maps produced using gradients of 
pre-softmax and post-softmax scores respectively.
As we can see, the relative performance of Grad-CAM++ and  Grad-CAM${}^+$
are very similar (relative performance $\approx 1$),
supporting our hypothesis that Grad-CAM++ is practically equivalent to Grad-CAM${}^+$.

\begin{table}[htb]
\caption{Relative performances of Grad-CAM methods across 
the ImageNetV2 dataset for images for which the network assigns 
a (post-softmax) score of more than $0.5$ to the right class.}\label{t:relperformance}
\begin{tabular}{|c|cc|}
\hline 
~ & \multicolumn{2}{c|}{\textbf{relative performance}} \\
\textbf{methods} & pre-softmax expl.\,maps & post-softmax expl.\,maps \\
\hline
\hline
Grad-CAM++ vs Grad-CAM & 1.24 & 1.27 \\
\hline 
Grad-CAM${}^+$ vs Grad-CAM & 1.16 & 1.25 \\
\hline
Grad-CAM++ vs Grad-CAM${}^+$ & \textbf{1.06} & \textbf{1.01} \\
\hline
\end{tabular}
\end{table}

We get also measures of dispersion from the 
distributions of log relative performances
per image across the dataset. We show the means and standard
deviations of the distributions in Table~\ref{t:logrelperformance}.

\begin{table}[htb]
\caption{Log relative performances of Grad-CAM methods across 
the ImageNetV2 dataset for images for which the network assigns 
a (post-softmax) score of more than $0.5$ to the right class.}\label{t:logrelperformance}
\begin{tabular}{|c|cccc|}
\hline 
 & \multicolumn{4}{c|}{\textbf{log relative performance}} \\
\textbf{methods} & 
\multicolumn{2}{c}{pre-softmax expl.\,maps} & 
\multicolumn{2}{c|}{post-softmax expl.\,maps} \\
~ & \phantom{----} mean & std & mean & std \\
\hline
\hline
Grad-CAM++ vs Grad-CAM & \phantom{----} 0.21 & 0.64 &  0.24 & 0.68 \\
\hline 
Grad-CAM${}^+$ vs Grad-CAM & \phantom{----} 0.15 & 0.66 & 0.22 & 0.68 \\
\hline
Grad-CAM++ vs Grad-CAM${}^+$ & \phantom{----} \textbf{0.06} & \textbf{0.41} & \textbf{0.01} & \textbf{0.11} \\
\hline
\end{tabular}
\end{table}

\begin{figure}[htb]
\begin{center}
\includegraphics[height=2in]{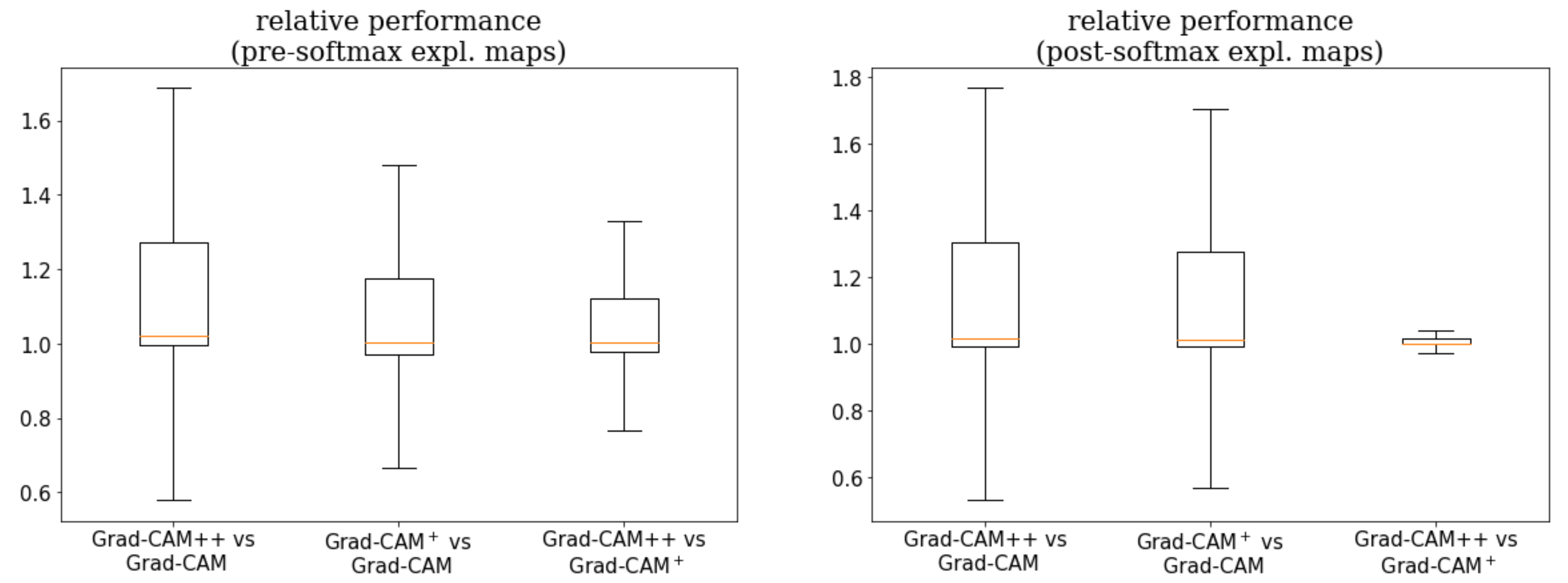}
\caption{Boxplots of distributions of 
relative performance per image.}\label{f:relperformance}
\end{center}
\end{figure}

Figure~\ref{f:relperformance} shows boxplots of relative performance
per image across the same dataset.  A value of~1 means same performance,
larger than~1 means the first method has better performance compared to the 
second, and less than~1 if the second does better than the first. 
We note that the distribution of the relative performance
again shows that Grad-CAM++ and Grad-CAM${}^+$ yield very similar results,
particularly if we use explanation maps obtained using gradients of post-softmax scores.

\section{Conclusions and Future Work}

We have examined the way Grad-CAM and Grad-CAM++ work, and compared them 
to a version of Grad-CAM, that we call Grad-CAM${}^+$, in which gradients
are replaced with positive gradients in the computation of the weights used
to combine the feature maps that produce heatmaps.  We have critically examined
the methodology behind the design of the Grad-CAM++ algorithm, uncovering a
number of weak points in it, and then showed that the algorithm is in fact
very approximately equivalent to the much simpler Grad-CAM${}^+$.

In future work additional effort can be dedicated to the comparison of
Grad-CAM++ vs Grad-CAM${}^+$ using different datasets and network models.
Also, although the methodology used to design the Grad-CAM++ is unclear,
the idea of a point-wise weighting (the alphas) during the computation of the 
weights ($w_k^c$) may still be salvaged, but equation (\ref{e:finalalpha2})
would need to be replaced with a new one that does not suffer from 
numerical instability and is properly justified.




\appendix

\section{Corrected Derivation for equation (\ref{e:firststep})}

In equation~(\ref{e:alphasequation}) we are assuming the following:

\begin{enumerate}
    \item $Y^c$ is a function of the $A_{ij}^{k}$.
    \item The $A_{ij}^{k}$ are independent variables not functionally related to each other,
hence $\frac{\partial A_{uv}^{l}}{\partial A_{ij}^{k}} = 1$ if $(i,j,k) = (u,v,l)$,
and $0$ otherwise.
    \item The $\alpha_{ij}^{kc}$ are treated as constants that do not depend on the $A_{ij}^{k}$,
hence $\frac{\partial \alpha_{uv}^{lc}}{\partial A_{ij}^{k}} = 0$.
\end{enumerate}

Also, following the Grad-CAM++ paper, we remove the $\text{ReLU}$ function.

Renaming summation indices as needed and computing the partial
derivative with respect to $A_{ij}^k$ of both sides of equation
(\ref{e:alphasequation}) (without $\text{ReLU}$) we get
\begin{equation}
  \begin{aligned}
    \frac{\partial Y^c}{\partial A_{ij}^k} &=
    \frac{\partial}{\partial A_{ij}^k}
    \sum_l \Bigl( \Bigl\{ \sum_{a,b} \alpha_{ab}^{lc} \cdot \frac{\partial Y^c}{\partial A_{ab}^l} \Bigr\}
    \Bigl[\sum_{u,v} A_{uv}^l\Bigr] \Bigr) \\
    &=
    \sum_l \frac{\partial}{\partial A_{ij}^k}  \Bigl( \Bigl\{ \sum_{a,b} \alpha_{ab}^{lc} \cdot \frac{\partial Y^c}{\partial A_{ab}^l} \Bigr\}
    \Bigl[\sum_{u,v} A_{uv}^l\Bigr] \Bigr) & \text{(par. deriv. inside sum)}\\
    & =
    \sum_l    \Bigl( \frac{\partial}{\partial A_{ij}^k}
    \Bigl\{ \sum_{a,b} \alpha_{ab}^{lc} \cdot \frac{\partial Y^c}{\partial A_{ab}^l} \Bigr\} \Bigr)
    \Bigl[\sum_{u,v} A_{uv}^l\Bigr] \\
    &\qquad +
    \sum_l \Bigl\{ \sum_{a,b} \alpha_{ab}^{lc} \cdot \frac{\partial Y^c}{\partial A_{ab}^l} \Bigr\}
    \Bigl( \frac{\partial}{\partial A_{ij}^k} \Bigl[\sum_{u,v} A_{uv}^l\Bigr] \Bigr) & \text{(product rule)}\\
    & =
    \sum_l
    \Bigl\{ \sum_{a,b} \alpha_{ab}^{lc} \cdot
    \frac{\partial}{\partial A_{ij}^k} \Bigl( \frac{\partial Y^c}{\partial A_{ab}^l} \Bigr) \Bigr\} 
    \Bigl[\sum_{u,v} A_{uv}^l\Bigr]  & \text{(par. deriv.  inside sum)} \\ 
    &\qquad +
    \sum_l \Bigl\{ \sum_{a,b} \alpha_{ab}^{lc} \cdot \frac{\partial Y^c}{\partial A_{ab}^l} \Bigr\}
    \Bigl[\sum_{u,v} \underbrace{ \frac{\partial A_{uv}^l}{\partial A_{ij}^k}}
    _{\substack{= 1 \text{ if } (u,v,l) = (i,j,k) \\ =0 \text{ otherwise} \hfill }}\Bigr]
                 & \text{(par. deriv.  inside sum)}\\   
    & =
    \sum_l \Bigl(
    \Bigl\{ \sum_{a,b} \alpha_{ab}^{lc} \cdot
    \frac{\partial^2 Y^c}{\partial A_{ij}^k \partial A_{ab}^l} \Bigr\} 
    \Bigl[\sum_{u,v} A_{uv}^l\Bigr] \Bigr) & \text{(combine par. deriv. )} \\
    &\qquad +
    \sum_{a,b} \alpha_{ab}^{kc} \cdot \frac{\partial Y^c}{\partial A_{ab}^k}  & \text{(simplify)} \\
    & =
    \sum_{a,b} \alpha_{ab}^{kc} \cdot \frac{\partial Y^c}{\partial A_{ab}^k} +
    \sum_l \Bigl(\Bigl[\sum_{u,v} A_{uv}^l \Bigr] 
    \Bigl\{ \sum_{a,b} \alpha_{ab}^{lc} \cdot
    \frac{\partial^2 Y^c}{\partial A_{ij}^k \partial A_{ab}^l} \Bigr\} \Bigr) 
    & \text{(rearrange terms)} \\  
  \end{aligned}
\end{equation}
We see that the result does not mach equation~(\ref{e:firststep}).

If we drop the assumption that the $\alpha_{ij}^{kc}$ are constant, then there
will be an extra term containing the partial derivatives of the alphas, and the
equation becomes:
\begin{equation}
\begin{aligned}
    \frac{\partial Y^c}{\partial A_{ij}^k} &= 
     \sum_{a,b} \alpha_{ab}^{kc} \cdot \frac{\partial Y^c}{\partial A_{ab}^k} +
    \sum_l \Bigl(\Bigl[\sum_{u,v} A_{uv}^l \Bigr] 
    \Bigl\{ \sum_{a,b} \alpha_{ab}^{lc} \cdot
    \frac{\partial^2 Y^c}{\partial A_{ij}^k \partial A_{ab}^l} \Bigr\} \Bigr) \\
    &\quad + 
    \sum_l \Bigl( \Bigl[\sum_{u,v} A_{uv}^l\Bigr] 
    \Bigl\{ \sum_{a,b} \frac{\partial \alpha_{ab}^{lc}}{\partial A_{ij}^k} \cdot \frac{\partial Y^c}{\partial A_{ab}^l} \Bigr\} \Bigr) 
    \,.
    \end{aligned}
\end{equation}

\end{document}